\documentclass[runningheads]{llncs}

\usepackage[T1]{fontenc}
\usepackage{graphicx}
\usepackage{todonotes}
\usepackage{hyperref}
\usepackage{graphicx}
\usepackage{booktabs}
\usepackage{subcaption}
\usepackage{siunitx}
\usepackage{float}
\usepackage{tabularx}
\usepackage{array}
\usepackage{enumitem}
\usepackage{rotating}
\usepackage{ragged2e}
\usepackage{amsmath}
\usepackage{amssymb}
\usepackage[font=scriptsize,skip=2pt]{caption}

\begin{document}
\title{SciEval: A Benchmark for Automatic Evaluation of K–12 Science Instructional Materials}
\titlerunning{SciEval: A Benchmark for Auto Evaluation for Instructional Materials}

% \author{
% Zhaohui Li\inst{1}\orcidID{0009-0002-0537-3546} \and
% Peng He\inst{2}\orcidID{0000-0002-2877-0117} \and
% Zhiyuan Chen\inst{2}\orcidID{0009-0007-0916-0595} \and
% Honglu Liu\inst{2,3}\orcidID{0009-0009-0897-5850} \and
% Zeyuan Wang\inst{2}\orcidID{0009-0002-5962-1277} \and
% Tingting Li\inst{2}\orcidID{0000-0002-5692-2042} \and
% Jinjun Xiong\inst{1,4}\orcidID{0000-0002-2620-4859}
% }
\author{
Zhaohui Li\inst{1} \and
Peng He\inst{2} \and
Zhiyuan Chen\inst{2} \and
Honglu Liu\inst{2,3} \and
Zeyuan Wang\inst{2} \and
Tingting Li\inst{2} \and
Jinjun Xiong\inst{1,4}
}

\authorrunning{Z. Li et al.}

\institute{
Department of Computer Science and Engineering, University at Buffalo, USA\\
\email{zli253@buffalo.edu}
\and
Department of Teaching and Learning, Washington State University, USA\\
\email{\{peng.he, zhiyuan.chen2, zeyuan.wang, tingting.li1\}@wsu.edu}
\and
College of Chemistry, Beijing Normal University, China\\
\email{honglu.alicia.liu@outlook.com}
\and
College of AI, Cyber and Computing, University of Texas at San Antonio, USA\\
\email{jinjun@utsa.edu}
}
\maketitle              % typeset the header of the contribution
\begin{abstract}

The need to evaluate instructional materials for K–12 science education has become increasingly important, as more educators use generative AI to create instructional materials.
However, the review of instructional materials is time-consuming, expertise-intensive, and difficult to scale, motivating interest in automated evaluation approaches. While large language models (LLMs) have shown strong performance on general evaluation tasks, their performance and reliability on instructional materials remain unclear. To address this gap, we formulate Automatic Instructional Materials Evaluation (AIME) as a generative AI task that predicts scores and evidence using the rubric designed by the educator. We create a benchmark dataset and develop baseline models for AIME. 
First, we curate the first AIME dataset, \textit{SciEval}, consisting of instructional materials annotated with pedagogy-aligned evaluation scores and evidence-based rationales. Expert annotations achieve high inter-rater reliability, resulting in a dataset of 273 lesson-level instructional materials evaluated across 13 criteria (N=3549) using the EQuIP rubric. 
Second, we test mainstream LLMs (GPT, Gemini, Llama, and Qwen) on \textit{SciEval} and find that none achieve strong performance.
Then we fine-tune \textit{Qwen3} on \textit{SciEval}. Results on a held-out test set show that domain-aligned fine-tuning can achieve up to 11\% performance gains, highlighting the importance of domain-specific fine-tuning for AIME and facilitating the use of LLMs in other educational tasks.

\end{abstract}
\section{Introduction}
 
High-quality instructional materials are central to achieving equitable, standards-aligned K-12 science education since they provide benefits to students' learning and teacher professional learning~\cite{Achieve2016a,BSCS2019,Carnegie2017,NASEM2018}. 
Recently, an increasing number of educators have begun using generative AI tools, such as ChatGPT, to create instructional materials, including lesson plans, assessment questions, and classroom activities. 
This phenomenon brings the need for automated evaluation of instructional materials, since manual evaluation is time-consuming and lacks reasoning details, with bias and errors, and limits scalability~\cite{li2024automate,camilli2024nlp,fu2025text}.
Recent research found that LLMs can be used for instructional material generation and evaluation, but LLMs struggle with pedagogical criteria, such as the depth of pedagogical reasoning and the coherence of instructional decisions \cite{hauk2025reliable,lee2024using}. 
In addition, most existing AI research focuses on content generation with human evaluation, but pays limited attention to systematically evaluating reliability and alignment with an official pedagogical rubric~\cite{zheng2025knowledge,tan2025elf,zheng2024automatic,ouyang2025lang}.

\begin{figure}[!t]
    \centering
    \includegraphics[width=\linewidth]{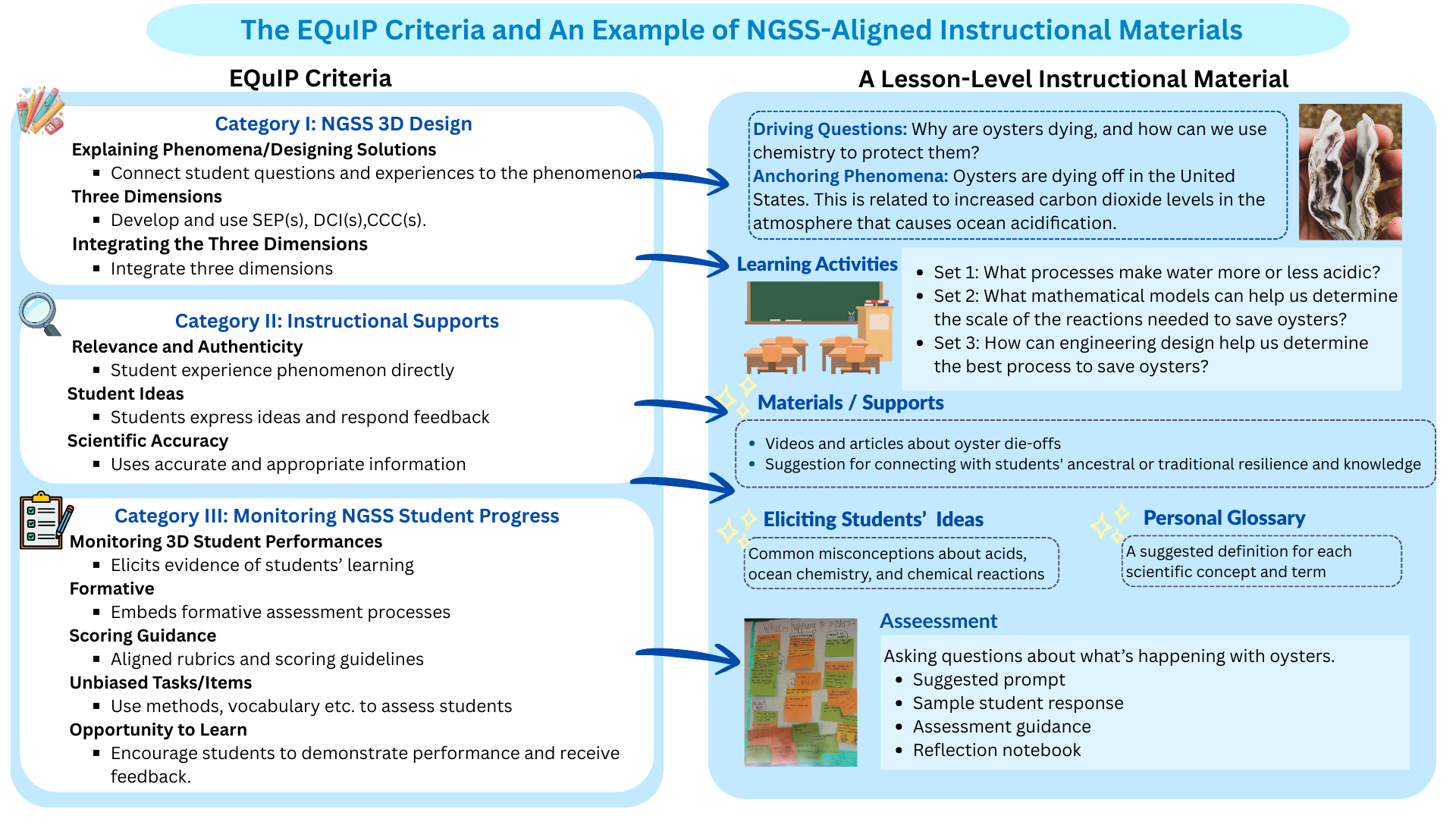}
    \caption{
    The left summarizes the EQuIP rubric criteria, and the right shows the corresponding lesson components, illustrating how rubric criteria align with instructional content for evaluation.}
    \label{fig:equip_mapping}
\end{figure}

Thus, we propose a new interdisciplinary task, Automatic Instructional Materials Evaluation (AIME), for both education and computer science communities. 
As illustrated in Figure \ref{fig:equip_mapping}, AIME aims to evaluate instructional materials by producing an overall quality score accompanied by evidence for every rubric dimension, which requires an algorithm to connect EQuIP~\cite{achieve2016equip} rubric criteria with the relevant content in instructional materials automatically. 
AIME benefits educators by reducing the assessment burden and has the potential to provide timely and detailed feedback. For AI research, AIME shifts the formulation of educational data evaluation from a classification problem that gives a score to a task that uses LLMs to generate feedback with evidence, which is more helpful and advanced in AI for Education~\cite{Yuan2024CourseEvaluationLLM,AbdulWahid2025CurriculumAlignedMCQ,Clark2025AutoEvaluation}. 
The key challenge of AIME is how to extract the evidence related to a rubric criterion in a long context.  Instructional materials are often presented in long PDF file, averaging 25 pages and about 6,000 words in our data. This causes the inherently long sequence modeling difficulty, which often leads to performance degradation in transformer-based LLMs~\cite{liu2024lost}. Furthermore, there is no reliable publicly available dataset for model training, particularly for LLM fine-tuning approaches that perform best on other educational tasks.

To address these challenges, we created the first AIME dataset for K-12 science education, named SciEval, from validated online open-source organizations, including OpenSciEd, BSCS, CREATE for STEM Institute, and the K-12 Alliance at WestEd with the NGSS Quality Badge.
Annotations were produced by two trained science education researchers using a multi-round training, calibration, and adjudication process guided by senior experts, resulting in high inter-rater reliability (89.7\% of agreement). The final dataset contains 273 lesson-level instructional materials, yielding 3,549 criterion-level annotation scores with evidence-based rationales. 
Building on SciEval, we conduct a three-step study to benchmark AIME task.
First, we co-designed and examined three structured prompts \ref{tab:prompts} with science education researchers. We found that the simplified prompt achieves the highest performance. 
Second, we benchmarked both commercial models and open-source instruction-tuned LLMs, and found that Qwen3-4B-Instruct \cite{yang2025qwen3} achieved the strongest performance.
Finally, we apply data augmentation on training data
and fine-tune Qwen3-4B-Instruct with LoRA-based supervised fine-tuning, and  techniques~\cite{dettmers2023qlora,hu2022lora}.
Results show fine-tuning improves performance on the score, while evidence grounding remains at the same level and motivates further work on long-context retrieval and evidence evaluation.

Our contributions are as follows. (1) We propose AIME, which evaluates instructional materials to support the safe and sustainable use of AI in education, and present SciEval, the first large-scale benchmark dataset for this task.
(2) We benchmark mainstream commercial and open-source LLMs on SciEval.
(3) We design a task-specific LLM fine-tuning approach with data augmentation for AIME, enabling stronger domain adaptation and improved performance that can be generalized to other educational tasks.

\vspace{-5pt}
\section{Related works}
\label{sec:related_works}
\vspace{-2pt}
Science educators in the United States have a strong consensus around \textit{A Framework for K–12 Science Education}~\cite{NRC2012} and the \textit{Next Generation Science Standards (NGSS)}~\cite{NGSS2013}. The EQuIP rubric provides structured criteria for assessing alignment to NGSS \cite{achieve2016equip}. Research has examined the design and implementation of NGSS-aligned materials and the practical challenges teachers face in using them effectively \cite{penuel2017rpp}. Studies have highlighted difficulties in applying these evaluation frameworks reliably across settings \cite{nordine2021promoting}. However, most of this work has focused on human-led processes and has not yet explored automated AI evaluation.

Several recent studies have explored LLMs for lesson plan research, but most focus on content creation rather than pedagogy-aligned evaluation.
EduGPT developed a generation model based on prompt engineering on top of GPT-4 and evaluated on three self-defined dimensions: Granularity, Accuracy, and Structure \cite{ye2024automatic}. 
ELF proposes a GPT Framework to generate teacher–student dialogical content to support classroom instruction \cite{tan2025elf}.  
Their evaluation relies on perplexity-based statistical metrics and human questionnaires on two self-collected datasets. 
LessonPlanLM \cite{zheng2025knowledge} defines three core features of high-quality lesson plans and fine-tunes a Qwen2-based model using over 100,000 self-collected plans. While this work advances large-scale lesson plan generation, the proposed features are not aligned with educational standards, and the dataset is not designed for automated evaluation. Similarly, LANG \cite{ouyang2025lang} collects 10,000 high-quality lesson plans and integrates RAG with LLMs to improve accuracy and reduce hallucinations. Those approaches emphasize generation quality rather than evaluation reliability.
Hauk's work~\cite{hauk2025reliable} reports that LLMs can approximate human judgments on some criteria, but struggle with higher-order pedagogical reasoning, such as instructional coherence and depth of learning design. 

From an AI perspective, prior datasets primarily contain high-quality lesson plans intended for generation, with few negative or borderline cases needed to train evaluation models. From an education perspective, unlike existing datasets that emphasize structural completeness or coarse holistic scores, our dataset provides fine-grained, NGSS-aligned, criterion-level annotations with explicit evidence grounding. This enables interpretable and standards-based evaluation rather than generation-oriented or surface-level scoring.
To the best of our knowledge, our work is the first to formulate AIME as a generative, evidence-grounded task aligned with educational standards.

\vspace{-10pt}
\section{Dataset}
\label{sec:dataset}
\vspace{-2pt}
\subsection{Data Source and EQuIP Evaluation Framework}
The dataset was constructed by integrating NGSS-aligned instructional materials, sourced from reputable organizations such as OpenSciEd, BSCS, CREATE for STEM Institute, and K-12 Alliance at WestEd, with corresponding evaluation reports conducted via the EQuIP rubric  
~\cite{NextGenScience2021Equip,achieve2016equip,NRC2012,NGSS2013}.
EQuIP was originally designed to support unit-level evaluation of instructional materials by providing criterion-level ratings and evidence-based rationales that reflect alignment with three-dimensional learning (Disciplinary Core Ideas [DCIs], Science and Engineering Practices [SEPs], and Crosscutting Concepts [CCCs], instructional supports, and assessment opportunities. DCIs refer to the core scientific ideas that students are expected to learn (e.g., energy, forces and motion, and related concepts), SEPs describe the practices scientists and engineers use to investigate phenomena and design solutions, and CCCs represent crosscutting concepts that apply across scientific domains (e.g., patterns, cause and effect, and structure and function) \cite{achieve2016equip}.

This study focuses on lesson-level evaluation, so we developed a systematic procedure to transfer unit-level EQuIP data to lesson-level annotations. Specifically, coders extracted criterion-level scores and associated rationales from official EQuIP reports and identified lesson-specific evidence supporting each criterion. When no lesson-level evidence could be traced for a given criterion, the corresponding lesson–criterion annotation was marked as missing data. This procedure preserved the evidentiary logic of the EQuIP framework while enabling fine-grained lesson-level analysis aligned with the goals of automated evaluation.

\subsection{Data Annotation}
Two science education researchers with strong familiarity with the NGSS and instructional materials served as coders. Two senior science education researchers led annotation training, calibration, and review processes. To ensure annotation consistency and validity, we follow a multi-round training and labeling process.

\textbf{Round 1: Training and Initial Reliability Assessment}. In Round 1, two instructional units comprising 29 lessons were selected. Two coders first reviewed the structure of NGSS-aligned lesson materials and the detailed definitions of each EQuIP criterion.  
They then independently annotated all lessons by extracting criterion-level scores and evidence-grounded rationales from EQuIP reports and organizing the annotations into structured tables. We used Cohen's kappa\cite{cohen1960coefficient} to calculate the inter-rater reliability (IRR). The initial IRR indicated fair to slight agreement: Cohen’s $\kappa$ = 0.227 (95\% CI: 0.12–0.34), with $\kappa$ = 0.327 for the high school unit and $\kappa$ = 0.131 for the middle school unit. Observed agreement was 60.6\% overall. Examination of the confusion matrix revealed systematic disagreement patterns, this indicates the difficulty of annotation. 
Following these results, the annotation team engaged in two structured adjudication meetings led by a senior researcher. These sessions focused on reviewing disagreement cases, clarifying decision boundaries, and resolving ambiguities in evidence attribution. Through this process, coders developed a shared set of decision rules and revised their initial annotations accordingly.

\textbf{Round 2: Validation and Reliability Improvement}. In Round 2 (Validation), two coders independently re-scored the same 29 lessons. In this round, IRRs were assessed using complementary metrics, including Cohen’s unweighted $\kappa$, linearly weighted $\kappa$, quadratically weighted $\kappa$, and the intraclass correlation coefficient ICC(2,1), with 95\% confidence intervals estimated via bootstrap resampling (1,000 iterations).
Post-calibration results demonstrated substantial improvement, with overall $\kappa$ = 0.724 (95\% CI: 0.65–0.78), ICC(2,1) = 0.725 (95\% CI: 0.65–0.78), and 89.7\% observed agreement.

\textbf{Round 3: Full Annotation}. Following calibration, coders proceeded to Round 3 (Annotation), coding the full dataset consisting of elementary (N = 8), middle school (N = 8), and high school (N = 7) instructional materials, totaling 23 instructional materials and 221 lessons. Coders independently annotated all lessons, with an overlapping 24 lessons double-coded to monitor reliability over time. The ICC(2,1) = 0.872, 95\% CI [0.83, 0.90]) with an overall Cohen's $k$ of 0.858, (95\% CI [0.81, 0.90]  and $95\%$ agreement indicate that the refined decision rules remained stable throughout the full annotation process.

\vspace{-5pt}
\subsection{Final Dataset}

The final dataset contains 273 lesson-level annotations, yielding 3,549 criterion-level instances with evidence-based rationales (see ground truth example in Table \ref{tab:case_hail_rain_ib1}). It spans 4,499 pages across 32 instructional units, with an average of 16.5 pages and 5,908 words per lesson. This scale supports fine-grained evaluation of LLMs on NGSS-aligned instructional materials, including both criterion-level scoring and evidence reasoning.
The dataset shows a pronounced long-tail distribution. Score~0 (N/A) accounts for 41.8\% (n=1{,}482) of all instances, while among the active scores (1–3), \emph{Extensive} performance (Score~3) dominates (54.2\%) and \emph{Inadequate} performance (Score~1) forms a small minority (6.5\%; n=135). The high proportion of Score~0 allows us to test whether models can appropriately abstain when evidence is missing. We split the data at the document level into training and test sets (218/55 PDFs) to prevent content leakage. To maintain comparable label distributions, we applied stratified sampling based on each PDF’s mean rubric score~\cite{singh1996stratified}. The dataset is publicly available on our website\footnote{\url{https://scieval-benchmark.github.io/SciEval/}}.

\vspace{-5pt}
\section{Methodology}
\label{sec:method}
\vspace{-5pt}

We conduct a three-step study to examine how LLM prompting could be applied to the AIME task. We first evaluate three different prompt designs. 
Then, we select and test candidate LLMs. Last, we fine-tune the best-performing model–prompt combination for domain adaptation.
\vspace{-5pt}
\subsection{Prompt Design}
\label{sec:prompt_design}
\vspace{-3pt}

We co-design three prompts with K-12 science education researchers to investigate how prompt affects model performance. Our design is motivated by prior findings on prompt sensitivity and hallucination in LLM research. First, adding domain knowledge in the prompt introduces a trade-off: while additional instructions may improve task specification, longer contexts can increase the likelihood of hallucination and lack of effective information~\cite{liu2024lost}.
Second, prompts containing dense professional terminology may overwhelm models, leading to degraded performance rather than improved reasoning~\cite{wei2022chain}.
Experiment results show that the simplified prompt has the best performance, as shown in Table \ref{tab:prompt_results}.
The three prompt designs, ranging from highly detailed to minimal and strictly constrained, are summarized in Table~\ref{tab:prompts}. \textit{Prompt A} is a Few-shot learning prompt, with professional terms like SEP and CCC, and evidence examples. \textit{Prompt B} is the same detailed prompt as A but zero-shot learning without examples. \textit{Prompt C} is the simplest prompt without professional terms.

\vspace{-2pt}
\begin{table}[!t]
    \centering
    \scriptsize
    \begin{tabular}{p{0.95\textwidth}}
        \toprule
        \textbf{Prompt A: Few-shot (Strong Domain Guidance + Examples)} \\
        \midrule
        \textbf{Design Idea:} Strong constraints, grounding rules, and many exemplars (highest token cost). \\

        \textbf{System Prompt (key parts):} \\
        Role: NGSS 3D learning activity evaluator. \\
        Inputs: parsed rubric + PDF text with page markers. \\
        Grounding: cite only marker page numbers; no invented quotes or pages; \\
        Output: JSON with \texttt{score} (0--3) and \texttt{evidence}. \\
        Evidence framing: explicitly reference NGSS dimensions (DCI / SEP / CCC + others). \\
        \textbf{User Prompt (schema + rules):} \\
        Schema: \texttt{\{"results":[\{"criterion\_id","score","evidence"\}]\}}. \\
        Constraints: exact keys only; \texttt{criterion\_id} must match the target; score $\in \{0,1,2,3\}$. \\
        Evidence: short summary + verbatim quotes with marker page numbers. \\
        Includes multiple score-2 / score-3 examples. \\
        \bottomrule

        \vspace{0.1pt}
        \textbf{Prompt B: Zero-shot (Reduced Domain Guidance)} \\
        \midrule
        \textbf{Design Idea:} Same structure as Prompt A, but without exemplars. \\

        \textbf{System Prompt (key parts):} \\
        Same as Prompt A except reference to DCI / SEP / CCC, but no explanation. \\
        \textbf{User Prompt (schema + rules):} \\
        Same schema and constraints as Prompt A. \\
        NGSS dimensions emphasized but not mandatory. \\
        No exemplar paragraphs. \\
        \bottomrule

        \vspace{0.1pt}
        \textbf{Prompt C: Simplified (No Domain Guidance)} \\
        \midrule
        \textbf{Design Idea:} Minimal rules to maximize compliance and reduce token cost. \\

        \textbf{System Prompt (key parts):} \\
        Role: NGSS-aligned instructional materials evaluator. \\
        Grounding (strict): no invented quotes or pages; verbatim only; marker pages only. \\
        Output (strict): JSON only; exactly ONE result; exact keys; score $\in \{0,1,2,3\}$. \\
        No NGSS dimension definitions (lowest token usage). \\

        \textbf{User Prompt (schema + rules):} \\
        Enforced format: \texttt{Summary: ... Evidence: "..." (Page X); "..." (Page Y)}. \\
        Constraints: JSON only; ONE item; exact keys; marker pages only. \\
        Evidence is a single string: 1--2 sentence summary + 1--3 short quotes. \\
        \bottomrule
    \end{tabular}
    \vspace{3pt}
    \caption{Three prompt designs for AIME: Few-shot, Zero-shot, and Simplified. We show simplified key components due to space limitations.}
    \label{tab:prompts}
    \vspace{-8pt}
\end{table}

\vspace{-10pt}
\subsection{Model Selection Rationale}
\label{sec:models}

Our model selection is guided by three principles.
First, we prefer models that demonstrate strong performance on educational benchmarks.
Second, we include both commercial and open-source models to enable comparison with AI systems commonly used by teachers (e.g., ChatGPT via GPT-4o-mini).
Third, we focus on small-scale models, rather than large foundational base models, because our dataset is relatively small and does not support full fine-tuning at the scale required by models with hundreds of billions of parameters. 
We exclude online chat tools such as ChatGPT and Claude from automated evaluation pipelines, as their outputs are not deterministic, cannot be systematically evaluated, and are not able to scale up in an AIME application.

In the first step, we selected \textit{Llama-3.1-8B-Instruct}, \textit{Llama-3.2-3B-Instruct}, and \textit{Qwen3-4B} (Instruct and Thinking), as they achieve SOTA performance among small LLMs on educational benchmarks such as MMLU and MATH, as shown in Table~\ref{tab:model-comparison}.
Massive Multitask Language Understanding (MMLU)~\cite{hendrycks2020measuring} evaluates academic knowledge across 57 subjects, including STEM domains.
The MATH benchmark~\cite{hendrycks2021measuring} measures symbolic and mathematical reasoning, which is essential for evaluating science instructional materials.
We also included \textit{GPT-4o-mini} and \textit{Gemini-2-Flash} because they are comparable in scale and performance to the selected open-source models and are widely used by educators through ChatGPT and Gemini chat.
Finally, for fine-tuning, we choose \textit{Qwen3-4B-Instruct-2507}, which achieves the strongest performance in the first-stage model selection (Table~\ref{tab:all_results}) and therefore provides a stronger foundation for domain adaptation.

\vspace{-8pt}
\begin{table}[H]
\centering
\scriptsize
\setlength{\tabcolsep}{4pt}
\renewcommand{\arraystretch}{1.15}

\begin{tabular}{lcccccc}
\toprule
\textbf{Model} & \textbf{Params} & \textbf{Context} 
& \textbf{MMLU} & \textbf{MATH} & \textbf{Access / Cost} \\
\midrule
GPT-4o-mini~\cite{hurst2024gpt}
& $\sim$8B & 128k  
& 82.0 & 70.2 
& API (\$0.15/\$0.60) \\

Gemini-2.0 Flash~\cite{comanici2025gemini}
& $\sim$17B & 1M  
& 88.4 & 72.0 
& API (\$0.30/\$2.50) \\

Llama-3.1-8B~\cite{dubey2024llama}
& 8B & 128k  
& 69.4 & 51.9 
& Open-source \\

Llama-3.2-3B~\cite{dubey2024llama}
& 3B & 128k  
& 63.4 & 48.0 
& Open-source  \\

Qwen-3-4B-Instruct~\cite{yang2025qwen3}
& 4B & 262k  
& 77.3$^{\dagger}$ & 54.1 
& Open-source \\

Qwen-3-4B-Think~\cite{yang2025qwen3}
& 4B & 262k  
& 83.7$^{\dagger}$ & 54.1 
& Open-source \\
\bottomrule
\end{tabular}

\vspace{0.3em}
\scriptsize
\textsuperscript{$\dagger$}Qwen3 reports \emph{MMLU-Redux} since classic MMLU is not available.  

\caption{Baseline LLMs and their educational reasoning performance. 
API cost is reported as \$/M input tokens \$/M output tokens.}
\vspace{-25pt}
\label{tab:model-comparison}
\end{table}

\vspace{-20pt}
\subsection{Fine-tuning Approach}

\begin{figure}[h]
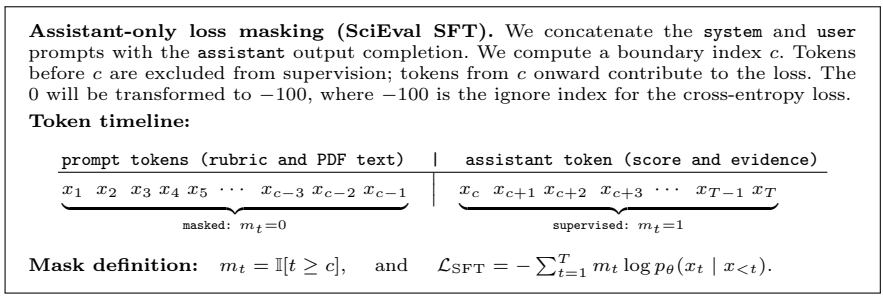

\centering
\scriptsize
\setlength{\fboxsep}{6pt}
\begin{minipage}{0.95\linewidth}
\fbox{
\begin{minipage}{0.95\linewidth}
\textbf{Assistant-only loss masking (SciEval SFT).}
We concatenate the \texttt{system} and \texttt{user} prompts
with the \texttt{assistant} output completion.
We compute a boundary index $c$. Tokens before $c$ are excluded from supervision; tokens from $c$ onward contribute to the loss. The $0$ will be transformed to $-100$, where $-100$ is the ignore index for the cross-entropy loss.

\vspace{2pt}
\noindent\textbf{Token timeline:}
\begin{center}
\ttfamily
\begin{tabular}{l}
\textbf{prompt tokens  (rubric and PDF text)} \hspace{2pt} | \hspace{2pt} \textbf{assistant token (score and evidence)} \\
\hline
\makebox[0.92\linewidth][l]{%
$\underbrace{x_1\;\;x_2\;\;x_3\;x_4\;x_5\;\cdots\;\;x_{c-3}\;x_{c-2}\;x_{c-1}}_{\text{masked: } m_t=0}
\;\;\; \Big|\;\;\;
\underbrace{x_c\;\;x_{c+1}\;x_{c+2}\;\;x_{c+3}\;\cdots\;\;x_{T-1}\;x_T}_{\text{supervised: } m_t=1}$} \\
\end{tabular}
\end{center}

\vspace{-2pt}
\noindent\textbf{Mask definition:}\quad
$m_t=\mathbb{I}[t \ge c]$,
\quad and \quad
$\mathcal{L}_{\text{SFT}}
= -\sum_{t=1}^{T} m_t \log p_\theta(x_t\mid x_{<t}).$
\end{minipage}
}
\end{minipage}
\caption{Assistant-only mask supervision used in SciEval fine-tuning.}
\vspace{-20pt}
\label{fig:masking}
\end{figure}

To establish stronger baselines for SciEval, we apply supervised fine-tuning (SFT) with Low-Rank Adaptation (LoRA) to fine-tune the \textit{Qwen3-4B-Instruct-2507} model. 
SFT enables the model to adapt domain knowledge in the SciEval training set \cite{ouyang2022training}. LoRA fine-tuning allows efficient fine-tuning with limited labeled data \cite{hu2022lora}. 
Specifically, we apply LoRA adapters to a subset of transformer projection layers \cite{hu2022lora}.
For the weight matrix
$W \in \mathbb{R}^{d \times k}$ in the pretrained weight in Qwen base model, LoRA defines the update as reparameterization:
\begin{equation}
W' = W + \Delta W, \qquad \Delta W = BA,
\end{equation}
where $A \in \mathbb{R}^{r \times k}$, $B \in \mathbb{R}^{d \times r}$ are the low-rank factors, and $r \ll \min(d,k)$ is the rank constraint. During training, $W$ is frozen and only $A$ and $B$ are updated.
We then follow the standard supervised fine-tuning (SFT) paradigm used in InstructGPT, where the training loss is masked on prompt tokens and computed only on the assistant completion~\cite{ouyang2022training}. 
Let $\mathbf{x} = (x_1,\ldots,x_T)$ denote the full token sequence, we follows a masking strategy as illustrated in Figure~\ref{fig:masking}. 
This masking strategy ensures that gradients are computed only from the assistant supervise information to prevent the model from memorizing rubric text and long instructional materials. Combined with LoRA, this allows the model to adapt to the SciEval domain efficiently, even with limited data and GPU memory.

\vspace{-15pt}

\vspace{-10pt}
\subsection{Label-Aware Resampling and Data Augmentation.}

To mitigate class imbalance during fine-tuning, particularly for the minority scores 1 and 2, we applied a label-aware resampling and translation-based data augmentation (DA) pipeline to the SciEval training set. Each instance is associated with a score $s \in \{0,1,2,3\}$.
We explored multiple resampling combinations, including downsampling the majority class and oversampling of minority classes. Pilot experiments show that the following sample method yields the best performance. 
First, we downsampled instances with $s=0$ from 1,181 (43.1\%) to 769 (33.0\%) to reduce majority-class dominance.
Second, we oversampled minority classes by following prior work that applies translation-based augmentation for imbalanced educational data~\cite{li2021semantic}. The number of $s=1$ instances was increased from 101 (3.7\%) to 303 (11.2\%), and $s=2$ from 585 (21.3\%) to 748 (27.7\%), while all original examples were preserved.
This procedure adjusts the training set from 2,743 to 2,696 instances while achieving a more balanced label distribution: 
28.5\% ($s=0$), 11.2\% ($s=1$), 27.7\% ($s=2$), and 32.5\% ($s=3$).

\vspace{-10pt}
\section{Experiments}
\label{sec:exp}

\vspace{-5pt}
\subsection{Experiment Settings}

The research questions our experiments address are: 1) How well does the current LLMs prompting method perform on SciEval? 2) Can domain-specific fine-tuning improve the model's performance? 
As introduced in section \ref{sec:models}, we test all models in Table \ref{tab:model-comparison} and a fine-tuned Qwen model. For each target criterion in SicEval testset, we extract and cache page-marked PDF text and prompt the model with EQuIP rubric context plus the full instructional material pdf text, requiring JSON-only outputs with score and grounded evidence
\footnote{Dataset and model code will be released if the paper is accepted.}.

We use 20\% of the training data as a development set for hyperparameter tuning. We test different combinations of hyperparameters and select the final configuration based on development performance. The chosen setting uses a learning rate $\eta = 3\times10^{-5}$, $E=6$ epochs, and LoRA rank $r=16$, with \texttt{batch\_size}=1 and gradient accumulation of 16 to control the effective batch size under long-context training (\texttt{max\_seq\_len}=8192, as each PDF is approximately 5{,}000 words). We fix \texttt{lora\_dropout}=0.05 and use a cosine learning rate schedule with \texttt{warmup\_ratio}=0.03. 
Both attention and feed-forward projection layers weight matrix are fine-tuned by LoRA, including \texttt{q\_proj}, \texttt{k\_proj}, \texttt{v\_proj}, \texttt{o\_proj},
\texttt{up\_proj}, \texttt{down\_proj}, and \texttt{gate\_proj}. 
We use deterministic decoding (\texttt{temperature}=0) with \texttt{max\_new\_tokens}=1024. Commercial models are evaluated using publicly available APIs with a temperature of 0. 
Each SFT fine-tuning run required approximately 10 hours, and inference on the full test set took about 4 hours on an NVIDIA H100 GPU.

\vspace{-10pt}
\subsection{Evaluation Metrics}
\label{sec:metrics}
\vspace{-5pt}

In AIME, models generate two outputs: a discrete \emph{score} and a free-text \emph{evidence} reasoning. For Score, we use Accuracy (Acc), Recall, QWK, and Macro-F1 (F1) as these traditional classification metrics that capture overall performance.
Evaluating generated evidence is more challenging, as no standard metric has been established. To address this challenge, we introduce an automatic, scalable metric inspired by natural language inference (NLI), named \emph{Evidence Match Rate (EMR)}. 
Let $E^{g} = \{e^{g}_{1}, \ldots, e^{g}_{m}\}$ and $E^{p}= \{e^{p}_{1}, \ldots, e^{p}_{n}\}$ represent the sets of ground-truth and predicted evidence sentences, respectively. For each $e_i^{p} \in E^{p}$, we compute its similarity to
all $e_j^{g} \in E^{g}$ using a semantic scoring function $s(\cdot,\cdot)$, implemented by sentence-embedding cosine similarity.
\vspace{-5pt}
\begin{equation}
\mathrm{EMR}
= \frac{1}{|E^{p}|}
\sum_{i=1}^{|E^{p}|}
\mathbb{I}\!\left[
\max_{e_j^{g} \in E^{g}} s(e_i^{p}, e_j^{g}) \ge \tau
\right].
\label{eq:emr}
\end{equation}

EMR measures the proportion of predicted evidence sentences whose maximum similarity to any ground-truth sentence exceeds a threshold $\tau$. We use the threshold $\tau = 0.65$ in experiments.

\vspace{-5pt}
\subsection{Results and Analysis}

\paragraph{Prompt selection}
We compared three prompting strategies, few-shot, zero-shot, and simplified prompt on both Llama-3.1-8B and Qwen-4b-Instruct models (Table~\ref{tab:prompt_results}). Notably, the simplified prompt C demonstrated robust performance across all metrics. For Llama, the simplified prompt C achieved the highest accuracy (43.80) and EMR (17.26), surpassing both few-shot and zero-shot variants. For Qwen, the simplified prompt also achieved the strongest overall performance, with the highest recall (31.64), F1 (29.50), and QWK (16.23). The results indicate that, for rubric-based instructional material evaluation, reducing prompt complexity enables the model to focus on the core task and mitigates overfitting. We chose this simplified prompt for all subsequent experiments.

\vspace{-10pt}
\begin{table}[H]
\centering
\scriptsize
\resizebox{\linewidth}{!}{
\begin{tabular}{l c c c c c}
\toprule
Model & Acc (\%) & Recall (\%) & F1 (\%) & QWK (\%) & EMR (\%) \\
\midrule

Few-shot Prompt (Llama)
& 33.29
& 24.16
& 20.62
& 6.84
& 12.18 \\

Zero-shot Prompt (Llama)
& 33.29 
& 24.16
& 20.62 
& 12.18 
& 14.97 \\

Simplified Prompt (Llama)
& {\bfseries 43.80 }
& {\bfseries 26.14 }
& 18.18
& 4.58
& {\bfseries 17.26} \\
\midrule

Few-shot Prompt (Qwen)
& 35.73 
& 28.78
& 23.37
& 11.18
& 37.56 \\

Zero-shot Prompt (Qwen)
& 33.86 
& 27.77
& 21.49
& 8.52 
& 35.02 \\

Simplified Prompt (Qwen)
& {\bfseries 38.90}
& {\bfseries 31.64}
& {\bfseries 29.50}
& {\bfseries 16.23}
& 36.04 \\

\bottomrule
\end{tabular}
}
\caption{Comparing the performance of different prompts on Llama and Qwen.}
\vspace{-30pt}
\label{tab:prompt_results}
\end{table}

\paragraph{Model selection.}

\begin{figure*}[h]
    \centering
    \includegraphics[width=\textwidth]{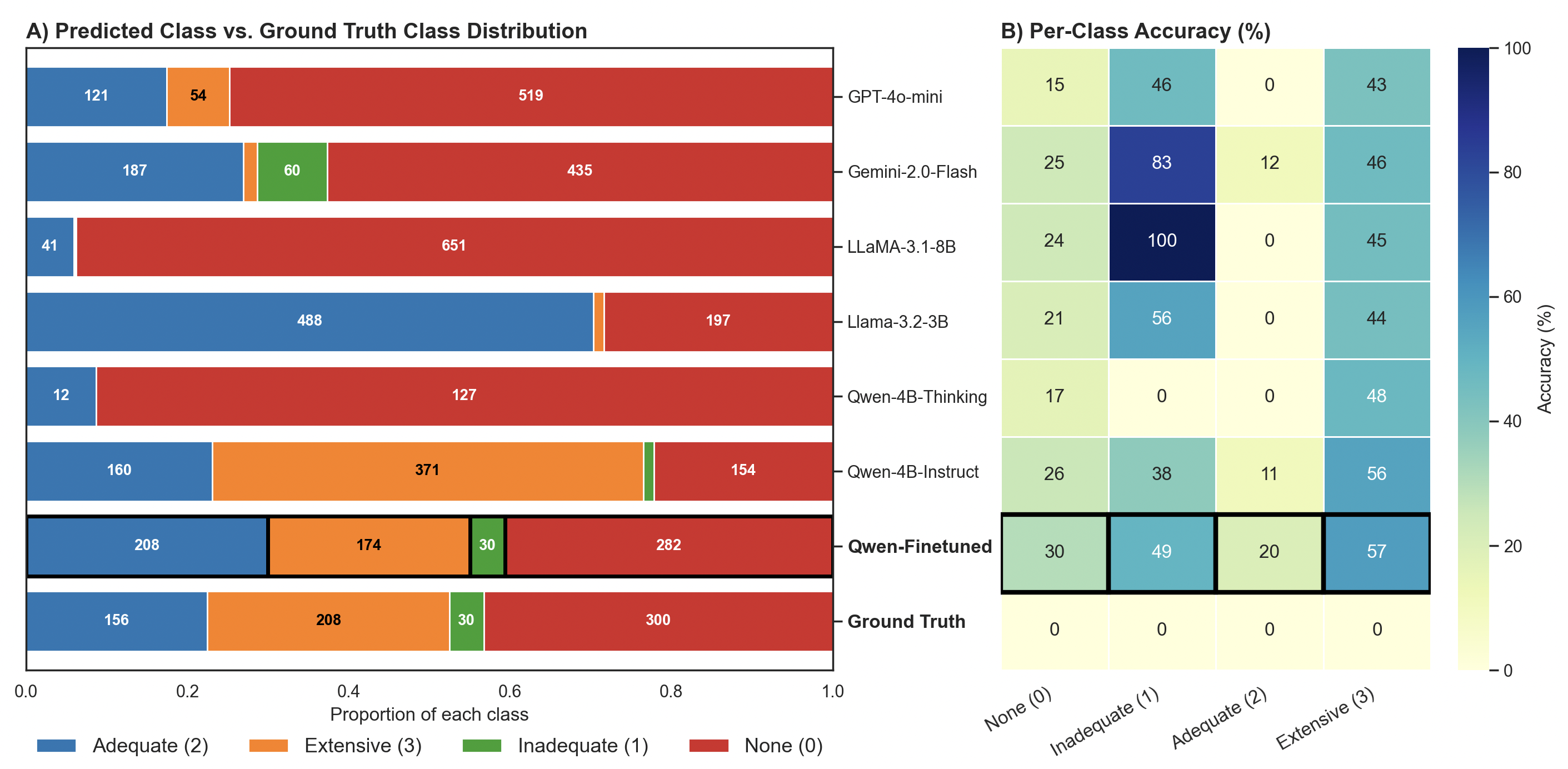}
    \caption{
    Distribution alignment and per-class accuracy across models on the SciEval test set.
    Note that the number of evaluated samples varies across models, as prediction failure cases are excluded. For example, the Qwen-4B-Think model produces a substantial number of incomplete outputs (e.g., label ``N''), resulting in smaller number.
    }
    \label{fig:dist_acc_jsd}
\end{figure*}

Among the base models, \emph{Qwen3-4B-Instruct} achieves the strongest overall balance, the highest Acc (38.90), F1 (29.50), QWK (16.23), and EMR (36.04), demonstrating superior agreement with human rubric scores and strong evidence finding. In contrast, \emph{Qwen3-4B-Thinking} gets the highest accuracy (53.85) but yields near-zero QWK and EMR, suggesting a collapse toward majority class predictions without meaningful rubric alignment. Therefore, we exclude this number from comparison. Commercial models such as GPT-4o-mini and Gemini-2.0-Flash achieve moderate accuracy but lower F1, QWK, and EMR, because they fail to predict 0 and 1 labels, as shown in Figure \ref{fig:dist_acc_jsd}.

\vspace{-15pt}
\begin{table}[H]
\centering
\scriptsize
\begin{tabular}{l c c c c c}
\toprule
Model & Acc (\%) & Recall (\%) & F1 (\%) & QWK (\%) & EMR (\%) \\
\midrule

GPT-4o-mini 
& 38.47 
& 24.56
& 21.70
& 4.02 
& 8.63 \\

Gemini-2.0-Flash 
& 38.04
& 31.16
& 26.54
& 5.70
& 18.27 \\

LLaMA-3.1-8B 
& 43.80 
& 26.14 
& 18.18 
& 4.58 
& 17.25 \\

LLaMA-3.2-3B 
& 27.95 
& 24.20
& 17.82 
& 4.52 
& 14.97 \\

Qwen3-4B-Instruct
& 38.90
& 31.64 
& 29.50
& 16.23
& {\bfseries 36.04} \\

Qwen3-4B-Thinking
& 53.85
& 25.00
& 17.50 
& 0.00 
& 0.00 \\

\midrule
Qwen3-4B-Instruct (Finetuned) 
& {\bfseries 49.28}
& 30.96
& 26.60
& 20.14
& 11.42 \\

Qwen3-4B-Instruct (Finetuned\&DA) 
&  45.53
& {\bfseries 38.77}
& {\bfseries 38.69}
& {\bfseries 25.73}
& 28.42 \\

\bottomrule
\end{tabular}
\caption{Comparing the performance of models on the test set from the SciEval.}
\vspace{-20pt}
\label{tab:all_results}
\end{table}

\paragraph{Fine-tuning results}

After applying LoRA-based supervised fine-tuning, both \emph{Qwen3-4B-Instruct}
(with or without DA) outperforms other baseline models in Acc. The DA model achieves the best Recall (38.77), F1 (38.69), and QWK (25.73). Compared to Qwen base model, fine-tuning with DA yields gains of 9.2\% in F1, 9.5\% in QWK, and 11.0\% in Recall. The DA has a big influence on the distribution of the predicted output, which leads to the increase of F1 and QWK.  Compared with other models, the fine-tuned model achieves up to 8\% across performance improvements in all metrics, demonstrating the effectiveness of domain-specific adaptation for the SciEval task. The fine-tuned model also achieves the closest distribution compared to ground truth while maintaining consistently high accuracy across all classes, as shown in Figure \ref{fig:dist_acc_jsd}.

In summary, the commercial models do not outperform the smaller open-source models, despite their larger parameter sizes and professional deployment. We find that the Instruct model outperforms the Think model on this task, and that fine-tuning with data augmentation improves overall performance. However, fine-tuned models remain limited in identifying and generating high-quality evidence. All models struggle to locate relevant evidence in long instructional texts, which remains a key challenge. We will discuss a representative example in the next section.

\vspace{-18pt}
\begin{table}[H]
\centering
\scriptsize
\setlength{\tabcolsep}{5pt}
\begin{tabular}{p{1.2cm} p{11.0cm}}
\toprule
\textbf{Source} & \textbf{Evidence Text} \\
\midrule

\textbf{Ground truth }(Score 3) &
``Analyzing and Interpreting Data. 
In Lesson 3, students analyze and interpret weather balloon data to determine how the temperature of the air higher up compares to the temperature of the air closer to the ground'' 
(Teacher Edition, \textcolor{red}{p.~88}). 
Element -- Use graphical displays (e.g., maps, charts, graphs, and/or tables) of large datasets to identify temporal and spatial relationships. \\
\midrule
\textbf{Base model }(Predict Score 3)  &
``Analyze and interpret weather balloon data to determine how the temperature of the air higher up compares to the temperature of the air closer to the ground'' 
(\textcolor{blue}{p.~2}); 
``Develop a model to show the relationship between the motion of the molecules that make up air and the energy of those molecules to explain the patterns of change in air temperature at various altitudes'' 
(\textcolor{green}{p.~1}); 
``The molecules that make up air high in the atmosphere have less kinetic energy, because it is cold. So we should draw those molecules a little closer together and moving more slowly. The molecules that make up air near the ground have more kinetic energy, because the temperature is higher. We should draw those molecules a little farther apart and moving more quickly'' 
(\textcolor{purple}{p.~12}). \\
\midrule
\textbf{Fine-tuned model } (Predict Score 3)&
``Analyze and Interpret Data to Identify Patterns in data in the graphs, charts, images, videos, maps, and/or text in order to describe patterns in the natural world in terms of the changes that occur in variables, controls, and constants that may have produced the patterns, including relationships governed by physical laws. 
Lesson 3: Say, Now that you have had the chance to look at data from two different sites, write a claim in your notebook that describes any relationships between the temperature of the air and the distance from the ground (altitude). Make sure you support your claim with evidence from your analysis of the data'' 
(Teacher Edition, \textcolor{orange}{p.~74}). \\

\bottomrule
\end{tabular}
\caption{Example from models output with ground truth for \textit{Hail\_rain\_lesson\_3} (Criterion \textit{I\_B\_1}).
Page numbers are color-coded to show evidence locations.}
\vspace{-20pt}
\label{tab:case_hail_rain_ib1}
\end{table}

\subsection{Discussion and Future Works}

We conducted an error analysis to better understand why fine-tuning did not consistently improve evidence quality. 
Table~\ref{tab:case_hail_rain_ib1} presents a representative case in which both the fine-tuned and base Qwen models predict the correct score. The fine-tuned model identifies the correct SEP and provides relevant supporting text, whereas the base model selects broader but less targeted evidence. However, the color coding shows neither model correctly locates the source page corresponding to the ground-truth evidence. This can't be reflected by EMR. It only measures the semantic level similarity. We need to conduct further human qualitative analysis on the evidence, especially for the page number. 
This failure reveals a limitation: even when the content is semantically appropriate, incorrect page localization prevents teachers from efficiently verifying and using the evidence. This motivates a future direction to jointly reason over content and precisely ground evidence to page-level locations, which is essential for actual teacher use.

Another limitation is that the SciEval training set is insufficient for fine-tuning larger models, such as Qwen3-Omni. This constraint limits our ability to investigate more advanced multimodal or long-context reasoning capabilities.
Additional concerns remain regarding overall performance, which is moderate for several reasons. First, the input sequence length is excessive due to the length of the PDF documents. In future work, we plan to identify relevant sections of the PDF for more targeted evaluation. We did not experiment across all settings, such as comparing three prompts across all models, due to time constraints. Inference on a 600-size test set can take up to 6 hours on an H100 GPU. We don't have enough computational resources. This is also why we did not fine-tune other models; we selected the best-performing Qwen. 

In future work, we plan to expand the SciEval dataset with additional expert-labeled examples and more detailed evidence annotations, including accurate page numbers. We will also explore new tasks such as suggestion generation and content creation. Finally, we aim to extend our framework to a broader range of teaching and learning settings.

\vspace{-8pt}
\section{Conclusion}

This paper presents a new task of Automatic Instructional Materials Evaluation, AIME, that aims to evaluate the quality of K–12 instructional materials by producing rubric-aligned scores together with pedagogical evidence. We collect a new interdisciplinary benchmark dataset to bridge LLM research and science education. 
We evaluate multiple mainstream LLMs on SciEval and establish a robust baseline by fine-tuning a Qwen-based model through supervised learning. Experimental results show that while current models achieve moderate performance in rubric-based scoring, they still struggle to extract reliable, precise evidence from long instructional documents. These findings highlight both the potential of LLMs for evaluating instructional materials and the need for future research on evidence grounding, long-context reasoning, and domain-specific alignment in educational applications.

\begin{credits}
\subsubsection{\ackname}This work is supported, in part, by the National Science Foundation under Grant 2229873 (AI4ExceptionalEd) and U.S. Department of Education under Grant R305C240046 (CELaRAI). This work was supported in part by the Boeing Distinguished Professorship at Washington State University (awarded to Tingting Li and Peng He). Any opinions, findings and conclusions or recommendations expressed in this material are those of the author(s) and do not necessarily reflect the views of the sponsors. 
\end{credits}

\begingroup
\tiny
\bibliographystyle{splncs04}
\bibliography{literatures}
\endgroup

\end{document}